\documentclass[conference]{IEEEtran}
\IEEEoverridecommandlockouts
\usepackage[nocompress]{cite}
 \usepackage[T1]{fontenc}
\usepackage{newtxtext}
\usepackage{caption}
\usepackage{amsmath,amssymb,amsthm}
\usepackage{graphicx}
\usepackage{booktabs}
\usepackage{multirow}
\usepackage[table]{xcolor}
\usepackage{url}
\definecolor{cvprblue}{rgb}{0.21,0.49,0.74}
\definecolor{citegreen}{HTML}{00B140}
\usepackage[colorlinks=true,citecolor=citegreen,linkcolor=cvprblue,urlcolor=cvprblue]{hyperref}

\newcommand{\B}[1]{\textbf{#1}}
\newcommand{\circled}[1]{\textcircled{\scriptsize #1}}
\newcommand{\dd}[1]{{\scriptsize\color{black!55}#1}}
\newcommand{\ddb}[1]{{\scriptsize\bfseries #1}}
\newcommand{\para}[1]{\noindent\textbf{#1}\hspace{0.4em}}
\definecolor{ours}{HTML}{EAF2FB}
\definecolor{greyrow}{HTML}{F2F2F2}

\DeclareMathOperator*{\argmax}{arg\,max}
\begin{document}
\bstctlcite{IEEEexample:BSTcontrol}
\title{Seeing Is Not Steering: Where Verifiable Reasoning Can Enter an End-to-End Driving Planner}
\author{\IEEEauthorblockN{%
Gautam Kumar Jain\textsuperscript{1},
Atharva Mahindrakar\textsuperscript{2,3},
Shubham Phoolari\textsuperscript{3},
Ignacio Alvarez\textsuperscript{2},
Carsten Markgraf\textsuperscript{1},\\
Constanze Hungar\textsuperscript{3},
Christoph Lauer\textsuperscript{3},
Julian Staehler\textsuperscript{1}
}
\IEEEauthorblockA{\textsuperscript{1}Technische Hochschule Augsburg/TTZ Landsberg, Germany  \quad \textsuperscript{2}Technische Hochschule Ingolstadt, Germany \\
\quad \textsuperscript{3}CARIAD SE, Germany}}

\maketitle

\begin{abstract}
Driving models that reason may describe a hazard correctly while choosing their path for other reasons. We study where scene reasoning can enter an end-to-end planner so its influence can be checked. We freeze Drive-JEPA and use a fine-tuned vision-language model to state scene facts. Although fine-tuning improves fact accuracy, the two learned components we test show no detectable benefit from reading them. A rule-based fact check instead rejects or shortens paths that contradict the facts, linking each intervention to a stated fact. On the frozen planner alone, it gains 2.4 EPDMS points on navhard. Its 2.0-point margin over a control that changes the same frames without consulting facts remains statistically uncertain. A diagnostic using exact facts from maps and annotations yields a 3.6-point margin over its corresponding control, with a 95\% confidence interval above zero. Separately, a path scorer and plan memory that read no stated facts raise navhard from 41.2 to 51.8. Bounding travel distance using the reasoner's facts adds 0.3 there and reduces navtest by 2.0. The fact check makes each intervention auditable, while the driving value of the reasoner's facts remains unresolved.
\end{abstract}

\section{Introduction}

Vision-language models (VLMs) can state useful facts about a driving scene, such as the traffic light, the vehicle ahead, or the free road. A driving planner, however, is judged by the path it drives, not by what a VLM says. A fact may be correct and still be ignored, while a trajectory may improve for reasons unrelated to that fact. We therefore ask whether the facts stated by a VLM actually influence a planner's decisions, and whether that influence improves driving.

This is hard to establish when reasoning is part of the policy itself. A fluent chain of thought need not be the reason for an answer~\cite{turpin2023unfaithful,lanham2023measuring}, vision-language-action models can show a gap between stated reasoning and action~\cite{foutter2026pinocchio}, and learning to reason does not necessarily improve driving~\cite{song2025disconnect}. A benchmark score alone also cannot reveal which fact, if any, caused a better trajectory.

We study this reasoning-to-action gap using Drive-JEPA~\cite{wang2026drivejepa}, an end-to-end planner that proposes and scores candidate paths. It performs strongly on navtest but drops substantially on navhard~\cite{cao2025pseudosim}, which also evaluates counterfactual states reached by the planner's own trajectories. We keep Drive-JEPA frozen so that any behavioral change comes from components added around it.

\begin{figure}[t]
\centering
\includegraphics[width=0.90\columnwidth]{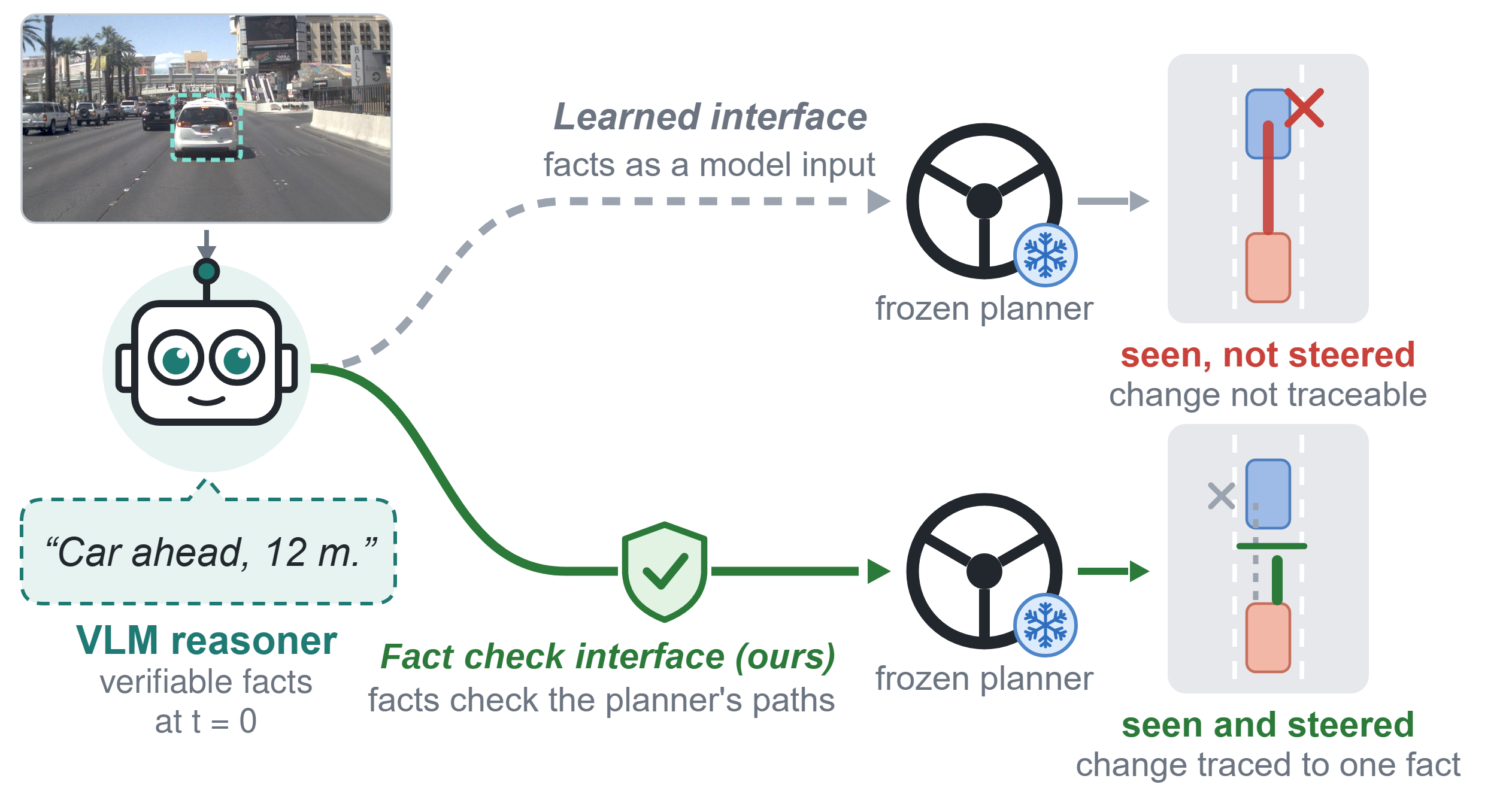}
\caption{\textbf{Seeing is not steering.} A VLM states checkable facts. Learned components (top) may change choices without revealing which fact mattered. Our explicit fact check (bottom) removes or shortens frozen-planner paths violating stated facts, making each intervention traceable.}
\label{fig:money}
\end{figure}

\begin{figure*}[t]
\centering
\includegraphics[width=0.88\textwidth]{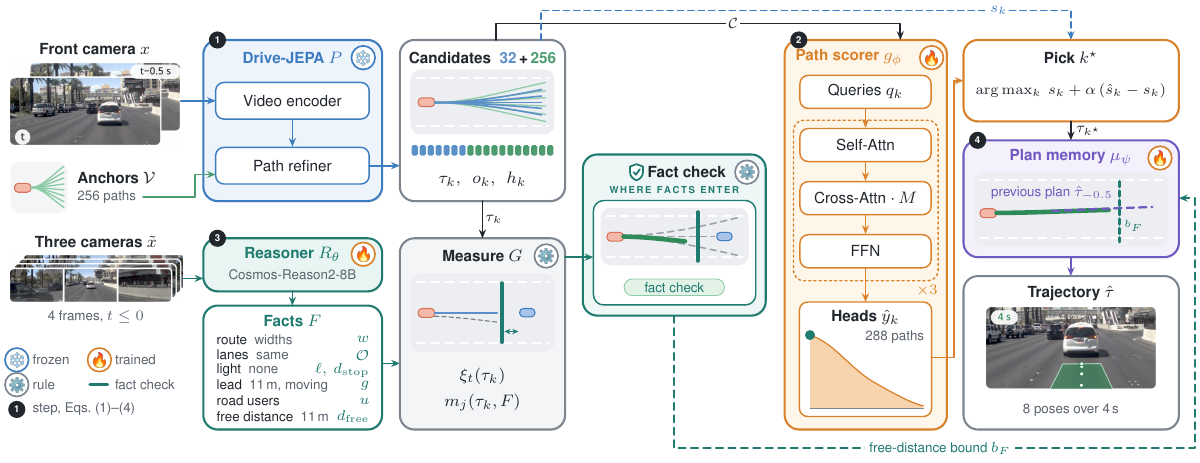}
\caption{\textbf{Architecture.} Frozen planner $P$ refines 32 paths and 256 anchors $\mathcal{V}$. Scorer $g_\phi$ predicts NAVSIM sub-scores; the pick combines them with planner score $s_k$; memory $\mu_\psi$ re-times the selected path toward the previous plan. Offline reasoner $R_\theta$ states facts $F$ at $t{=}0$. The fact check filters frozen-planner paths using five clauses or bounds full-system progress by $b_F$. Circled numbers denote Eqs.~(\ref{eq:propose})--(\ref{eq:check}), not rows of Table~\ref{tab:steps}.}
\label{fig:arch}
\end{figure*}

Our approach separates seeing from steering (Fig.~\ref{fig:money}). An offline Cosmos-Reason2-8B reasoner~\cite{nvidia2025cosmosreason}, fine-tuned for driving, writes a structured \emph{fact sheet} describing scene attributes such as the traffic light and stop line, vehicle ahead, free distance, adjacent lanes, and surrounding road users. We test two interfaces: learned components can consume these facts as inputs, or an explicit \emph{fact check} can remove or shorten candidate paths that violate them.

The fact check makes every intervention traceable to a stated fact. To test whether the gain comes from factual content rather than from changing the path itself, we compare against shuffled facts from another scene and a matched control that changes the same frames without using a fact.

We also build a stronger base system around the frozen planner using a learned path scorer and plan memory, neither of which reads the facts. The full system then adds the reasoner's free-distance constraint. This lets us ask not only whether reasoning helps, but where: on scenes where the frozen planner already fails, or on rare scene types.

Our contributions are:

\begin{itemize}[\IEEEnocalcleftmargintrue\setlength{\leftmargin}{0pt}\setlength{\itemindent}{\dimexpr\labelwidth+\labelsep\relax}]
\item \textbf{A controlled test of stated facts.} We test whether facts are correct, whether they change the driven path, and whether that change improves the score, using shuffled and matched controls.

\item \textbf{A verifiable reasoning-to-action interface.} On navhard, the explicit fact check scores 2.4 EPDMS above no check and 3.5 above shuffled facts. Its 2.0-point margin over the matched control includes zero. With exact facts, used only as a diagnostic, the margin is 3.6 points. The two learned components we tried did not benefit from the same facts.

\item \textbf{A stronger frozen-planner system.} The path scorer and plan memory raise Drive-JEPA from 41.2 to 51.8 EPDMS on navhard. Adding the free-distance bound gives 52.2 on navhard, within noise of the base system, and lowers navtest from 90.2 to 88.2.

\item \textbf{Hard, not rare.} The free-distance bound helps mainly where the frozen planner already fails and hurts where it does not. We find no additional benefit on rare scene types, where the reasoner's facts are also less accurate.
\end{itemize}

\noindent These results separate \emph{reading reasoning} from \emph{acting on it}: explicit constraints make reasoning interventions auditable, but their benefit depends on where and how they are applied.

\section{Related Work}

\para{Reasoning inside driving policies.} A VLM either answers structured questions and emits a decision that a planner follows, as in DriveLM~\cite{sima2024drivelm} and DriveVLM~\cite{tian2024drivevlm}, or produces reasoning and trajectory together, as in Alpamayo-R1~\cite{nvidia2025alpamayo} and ELF-VLA~\cite{luo2026elfvla}. These systems show that language reasoning can be connected to driving decisions, but the reasoning is internal to the driving policy and is ultimately judged through its driving score. This makes it difficult to separate whether performance comes from the stated reasoning itself or from other representations learned jointly with the action. We instead keep the reasoner outside a frozen planner, restrict its output to scene facts that can be checked independently, and measure how those facts change the planner's candidate paths. Safety layers~\cite{shalev2017rss,alshiekh2018shield} also constrain a policy from outside, but their rules are hand-written rather than produced from what a model states about the current scene. Our setting therefore lies between learned reasoning and external safety constraints: the constraint is explicit, but its content comes from the VLM.

\begin{figure*}[t]
\centering
\includegraphics[width=0.78\textwidth]{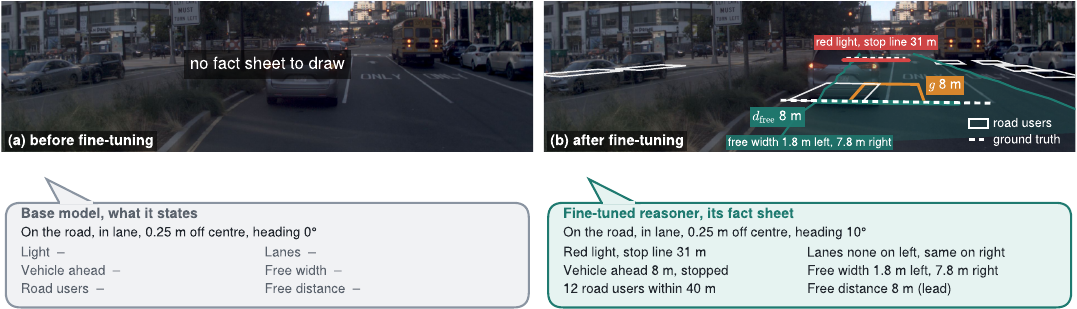}
\caption{\textbf{Seeing, before and after fine-tuning.} The same held-out real scene at $t{=}0$. (a)~Base Cosmos-Reason2-8B, with the same prompt and token budget, states only the car's location before looping on numbers; dashes mark unstated facts. (b)~The fine-tuned fact sheet overlaid on the camera: teal for free widths and distance $d_{\mathrm{free}}$, orange for lead-vehicle gap $g$, red for light and stop line, white for road users. Dashed white shows ground truth.}
\label{fig:facts}
\end{figure*}

\para{Faithfulness and reliance.} An explanation is faithful if it reflects the process behind the decision~\cite{jacovi2020faithful}. Prior work shows why this cannot be assumed from language alone: perturbing a chain of thought often leaves the answer unchanged~\cite{turpin2023unfaithful,lanham2023measuring}, and in driving, learned reasoning does not always shape the resulting plan~\cite{song2025disconnect}. Inside Alpamayo-R1, small image perturbations can alter actions up to collisions while rationales remain fixed~\cite{feng2026saysame}, showing that stated reasoning and action may diverge even within the same model. More generally, giving a model an informative input does not establish that its decision relies on that input rather than on a correlated signal~\cite{dehaan2019causal,li2024egostatus}. Our question is therefore not only whether the reasoner states the right facts, but whether those facts are actually used. We check each stated fact against the scene, test whether learned components respond to it, and, for the explicit fact check, trace each intervention back to the statement that caused it.

\para{Trajectory scoring and temporal consistency.} A separate line of work improves driving by selecting better trajectories rather than by adding language reasoning. PDM-Closed~\cite{dauner2023pdm} selects rule-based proposals using privileged inputs, while Hydra-MDP~\cite{li2024hydramdp,li2025hydramdppp}, GTRS~\cite{li2025gtrs} and DrivoR~\cite{kirby2026drivor} distil simulator scores into networks that rank proposals or fixed anchor paths. Other planners improve consistency between consecutive plans~\cite{song2025momad,li2025ztrs,li2025hydramdppp,pei2026world4scorer}. Our path scorer and plan memory follow these directions to improve the frozen Drive-JEPA planner without using the reasoner's facts. This gives us a stronger non-reasoning baseline on which we can test whether the facts provide additional value, while keeping their effect separate from trajectory scoring and temporal consistency.

\section{Method}\label{sec:method}
We now describe the system used to test how stated facts can affect a frozen driving planner. The method separates the planner, the reasoner that produces the fact sheet and the components through which those facts can influence the driven path. We then build a stronger system around the same frozen planner to test whether the facts still add value when the planner is already improved.
\subsection{Overview}\label{sec:overview}
Camera images up to the present, $t{=}0$, go in, and one path of eight poses over the next 4\,s comes out and is scored by NAVSIM. The system has four steps, shown in Fig.~\ref{fig:arch},
\begin{align}
\mathcal{C} &= P(x,\,\mathcal{V}),\label{eq:propose}\\
\{\hat{y}_k\}_{k=1}^{N} &= g_\phi(\mathcal{C}),\label{eq:score}\\
F &= R_\theta(\tilde{x}),\label{eq:see}\\
\hat{\tau} &= \Pi_F\big(\mu_\psi(\tau_{k^\star},\,\hat{\tau}_{-0.5})\big).\label{eq:check}
\end{align}
The frozen planner $P$ proposes candidates $\mathcal{C}$ from the front camera $x$ and the anchors $\mathcal{V}$, 256 fixed paths (Sec.~\ref{sec:candidates}). The path scorer $g_\phi$ predicts each candidate's simulator sub-scores $\hat{y}_k$, from which the pick selects $k^\star$ (Sec.~\ref{sec:scorer}). The plan memory $\mu_\psi$ re-times the chosen path toward the previous plan $\hat{\tau}_{-0.5}$ (Sec.~\ref{sec:memory}). The reasoner $R_\theta$ reads three cameras $\tilde{x}$ offline and writes a fact sheet $F$ (Sec.~\ref{sec:see}), and the fact check $\Pi_F$ bounds how far the re-timed path goes (Sec.~\ref{sec:facts}). The planner, scorer and memory form the \emph{base system}, and the \emph{full system} adds the fact check. On the frozen planner alone, the fact check instead filters the candidates.

Sec.~\ref{sec:candidates} to \ref{sec:attribution} cover the study of the facts on the frozen planner, and Sec.~\ref{sec:scorer} and \ref{sec:memory} cover the scorer and the memory. One rule holds throughout. The reasoner's inputs end at $t{=}0$, and every fact is compared with a label computed at $t{=}0$, never from the future.

\subsection{Candidates from the frozen planner}\label{sec:candidates}
Drive-JEPA~\cite{wang2026drivejepa} encodes the current and the previous front-camera image, $x$, with its pretrained video encoder and refines $K{=}32$ path queries against the image tokens. Each query becomes a \emph{candidate path} $\tau_k\in\mathbb{R}^{8\times3}$ of eight poses 0.5\,s apart, with six score logits $o_k\in\mathbb{R}^{6}$ and eight waypoint tokens $h_k\in\mathbb{R}^{8\times256}$. The logit of its final score is $s_k$. Left alone, the planner drives $\pi=\argmax_{k\le K}s_k$. Its paths stay near its own preference, so we add 256 \emph{anchors}, the $k$-means centres of its paths on training scenes, as score-based planners do~\cite{li2024hydramdp,li2025gtrs}. Each anchor takes the slot of a path query in the frozen refiner and receives logits and tokens like a planner path. The candidate set is $\mathcal{C}=\{(\tau_k,o_k,h_k)\}_{k=1}^{N}$ with $N{=}288$, and $\mathcal{V}$ also denotes the anchor indices $k>K$.
\subsection{Facts from the reasoner}\label{sec:see}
The reasoner $R_\theta$, Cosmos-Reason2-8B~\cite{nvidia2025cosmosreason} with a low-rank adapter (LoRA)~\cite{hu2022lora}, runs offline on $\tilde{x}$. This input holds the three front cameras at four frames from $t{=}{-}1.5$\,s to $0$ and the ego history. The reasoner writes the \emph{fact sheet} $F$ as structured text about the present scene. The sheet lists the free width $w$ to the drivable edge on each side of the route, the adjacent lanes and which of them are oncoming ($\mathcal{O}$), the light $\ell$ and the distance $d_{\mathrm{stop}}$ to its stop line, the gap $g$ to the vehicle ahead, and the positions and speeds of the road users $u$. It also lists the \emph{free distance} $d_{\mathrm{free}}$ to the first required stop. Simulator verdicts and future positions are not part of it.

\subsection{The fact check}\label{sec:facts}
\para{Measure.} The simulator scores the path that the controller drives, so a fixed rule $G$ tracks each candidate with the benchmark's controller and bicycle model~\cite{dauner2024navsim}. It compares the poses $\xi_t(\tau_k)$ with the sheet through five measures $m_j$. The margin $m_{\mathrm{corr}}$ is how far the car's worst corner stays inside the stated free width. The clearance $m_{\mathrm{clr}}$ is the distance to a stated road user moved at its stated velocity. The distance $m_{\mathrm{onc}}$ is driven in an oncoming lane, and the stop margin $m_{\mathrm{stop}}$ is the space before a red light's stop line. The overrun $m_{\mathrm{free}}$ is how far the front bumper goes beyond $d_{\mathrm{free}}$. A path breaks a fact if any of five clauses fails,
\begin{equation}
\begin{gathered}
m_{\mathrm{corr}}\ge-0.5\,\mathrm{m},\quad m_{\mathrm{clr}}\ge0.5\,\mathrm{m},\quad m_{\mathrm{onc}}\le0,\\
m_{\mathrm{stop}}\ge0,\quad m_{\mathrm{free}}\le3\,\mathrm{m},
\end{gathered}
\label{eq:clauses}
\end{equation}
with tolerances set on SimScale critical scenes (Sec.~\ref{sec:splits}) and never on the test splits. We write $c_j(\tau,F)$ for the statement that clause $j$ holds.

\para{Two forms of the fact check.} The \emph{five-clause fact check} runs on the frozen planner, where Sec.~\ref{sec:notwork} studies the facts. It removes the paths that break a clause and leaves the choice to the planner,
\begin{equation}
\mathcal{S}(F)=\{k\le K\mid c_j(\tau_k,F)\,\forall j\},\quad \pi_F=\argmax_{k\in\mathcal{S}(F)} s_k ,
\label{eq:filter}
\end{equation}
or $\pi$ if no path survives, so it adds no preference of its own. The \emph{free-distance bound} runs in the full system. Only the free-distance clause acts, as a bound on the progress that the plan memory returns,
\begin{equation}
\lambda_i'=\min\big(\tilde{\lambda}_i,\ b_F\big),\qquad b_F=d_{\mathrm{free}}+3\,\mathrm{m},
\label{eq:bound}
\end{equation}
and a frame without a fact sheet has $b_F{=}\infty$. Both forms change the driving only where a stated fact is broken. Every change therefore has a cause that we can check, whether the fact is right or wrong.

\para{Learned components.} For comparison, two learned components read the same facts as inputs. A \emph{re-ranker} scores the planner's 32 paths, is trained on simulator outcomes and reads the fact sheet and each path's five measures. A \emph{retrained head} is the planner's head, retrained with the facts as input and with one objective, the distance by which its paths break them.

\subsection{Evaluating stated facts}\label{sec:attribution}
A fact matters only if it is true and it changes what the car does, and a benchmark score alone cannot separate the two. We therefore ask three questions of every stated fact, from the fact itself to its effect on driving.

\para{Right. Is it true?} Each fact is compared with a label computed from the map and the agents at $t{=}0$. A category must match, the free distance must lie within 2\,m, and a gap or width within 1\,m or 10\%, whichever is larger (Fig.~\ref{fig:facts}, Table~\ref{tab:right}).

\para{Used. Does it change the driving?} The fact check \emph{changes} a frame when the path it drives differs from the path driven without it. A change \emph{fixes} a hard failure (a collision, leaving the drivable area, driving against traffic or running a red light) when the path without the fact check fails and the path with it does not, and \emph{introduces} one in the reverse case.

\para{Helps. Does the score rise?} We compare paired EPDMS point changes against the shuffled and matched controls of the introduction. A learned component that reads the facts can be compared with the same controls, but its changes have no stated cause (Sec.~\ref{sec:notwork}).

\subsection{Path scorer}\label{sec:scorer}
The planner's own scores choose poorly among acceptable paths, so the path scorer predicts, for every candidate, the nine sub-scores the simulator would give it~\cite{dauner2024navsim}. Each candidate becomes a query of width 256, $q_k=\mathrm{MLP}_g([\tau_k,o_k])+\mathrm{MLP}_h(\bar{h}_k)$, where $\bar{h}_k$ joins the maximum and the mean of its waypoint tokens. Three transformer blocks (3.7\,M parameters) let the candidates attend to each other and to a context of all candidates' pooled tokens, an image token and an ego token. A linear head gives $\hat{y}_k\in[0,1]^9$. Four scorers, trained with different data and seeds, are averaged, and $\hat{s}_k$ is their mean final-score logit. An anchor has no planner score, so we set $s_k{=}\hat{s}_k$ for it and let it win only by a margin $\delta$. The pick mixes scorer and planner, $v_k=\sigma\big(s_k+\alpha(\hat{s}_k-s_k)\big)-\delta\,\mathbf{1}[k\in\mathcal{V}]$, with logistic $\sigma$, $\alpha{=}0.7$ and $\delta{=}0.05$. Of the planner's paths within $\epsilon{=}0.02$ of the best value, the one with least jerk and yaw acceleration, $k_P$, is kept, and $k^\star=\argmax_{k\in\{k_P\}\cup\mathcal{V}}v_k$. With $\alpha{=}\epsilon{=}0$ and no anchors, $k^\star{=}\pi$.

\subsection{Plan memory}\label{sec:memory}
A system that makes a fresh choice in every frame may switch between acceptable paths with different speeds. The plan memory keeps the chosen path's geometry and re-times it to continue the previous plan $\hat{\tau}_{-0.5}$, which is recomputed by running the system on the frame's own camera history. With $\lambda_i$ the distance along the path to pose $i$ and $L{=}\lambda_8$, the path is first re-timed so that its acceleration continues the present value $a_0$, $a(t)=a_\infty+(a_0-a_\infty)e^{-t/0.6\,\mathrm{s}}$, with progress $\min(L,b_F)$ at 4\,s, where $b_F$ is the bound set by the facts (Sec.~\ref{sec:facts}). A multilayer perceptron (MLP) $\mu_\psi$ then reads $z$, which holds the path, its progress increments $\Delta\lambda_i$, the ego state, the motion of the current and the previous plan over their 3.6\,s overlap and its distance to the comfort limits. The first of its eight heads proposes changes $\delta=2.5\tanh(\mu_\psi(z)_1/2.5)\in\mathbb{R}^8$ to the logs of the increments, so that $\tilde{\lambda}_i=\sum_{j\le i}\exp(\log\Delta\lambda_j+\delta_j)$.

\section{Experimental Setup}

\subsection{Data}\label{sec:splits}
We use NAVSIM v2~\cite{dauner2024navsim,cao2025pseudosim}. Training uses navtrain, real scenes with recorded camera images, and SimScale~\cite{tian2026simscale}, critical scenes simulated from navtrain with rendered camera images. We test on navtest, 12{,}146 real scenes of which 10{,}795 have camera files, and on navhard, 225 groups of real frames and the rendered start states simulated from them (5{,}912 frames). The official navtest scores of Tables~\ref{tab:main} and \ref{tab:steps} cover all 12{,}146 scenes, and the analyses of the fact check and of the slices use the 10{,}795 scenes with camera files.

\subsection{Training}\label{sec:training}
The planner is the published NAVSIM Drive-JEPA~\cite{wang2026drivejepa} and stays frozen. It omits its released momentum step, since that step applies the benchmark's two-frame comfort check in the simulator at test time. With the step it scores 90.8 on navtest (Table~\ref{tab:steps}). The reasoner, path scorer and plan memory are trained offline, once before any test run, on training scenes only. None is tuned on navtest or navhard.

\para{Reasoner.} The reasoner is fine-tuned on 9{,}851 real navtrain scenes and 11{,}482 rendered SimScale scenes, on fact sheets $y$ whose values are computed from the map and the agents at $t{=}0$. A single wrong value can decide a critical scene while most of the sheet is boilerplate, so the loss weights each token by its role,
\begin{equation}
\mathcal{L}_{\mathrm{R}} = -\frac{\sum_j \omega_j \log p_\theta\big(y_j \mid y_{<j}, \tilde{x}\big)}{\sum_j \omega_j} + \lambda_{\mathrm{num}}\, \mathcal{L}_{\mathrm{num}},
\label{eq:lreason}
\end{equation}
with $\omega_j{=}3$ for decision values (light, vehicle ahead, free distance, lanes), 1 for other values and 0.1 for syntax, and a number-token loss $\mathcal{L}_{\mathrm{num}}$~\cite{zausinger2025ntl} that grows with the numeric error ($\lambda_{\mathrm{num}}{=}0.3$). Its last round, the \emph{repair}, oversamples red lights near a stop line and slow vehicles 20 to 40\,m ahead, the two most frequent errors before it. The studies of the five-clause fact check use the reasoner before the repair, and the full system uses the repaired one. The base system does not read the facts. Table~\ref{tab:right} evaluates the reasoner before the repair, on 1{,}600 held-out scenes, 800 real and 800 rendered, and Sec.~\ref{sec:reasoner} gives the repair's changes.

\para{Path scorer.} Four scorers are trained on 11{,}807 navtrain scenes plus a SimScale subset of 47{,}279 or 97{,}888 scenes, with two seeds per subset size, and SimScale validation scenes select the epoch. Their labels are the nine simulator sub-scores of each candidate, used as targets and never as inputs. Each scorer minimizes a binary cross-entropy over them, weighted 2 for the final score and 1 for the others.

\para{Plan memory.} The memory learns from 180{,}273 pairs of consecutive frames in SimScale scenes. For each pair, a search drives re-timings of the current plan through the benchmark's controller and keeps those that bring the two plans within its comfort limits~\cite{cao2025pseudosim}. The eight that change the plan least are the targets of the eight heads, and the first head, the only one used at test time, is also pulled toward the smallest compliant change. Training alternates between this search, seeded with the network's own proposals, and fitting the heads.

\subsection{Metric and protocol}\label{sec:metric}
We report the extended predictive driver model score (EPDMS) of NAVSIM v2~\cite{cao2025pseudosim}. Hard checks for collisions, drivable area, driving direction and traffic lights gate a weighted average of progress, time to collision, lane keeping, and comfort within and between plans. On navhard, each real frame's score is multiplied by the average score of its simulated follow-ups, weighted by their distance to the plan's end. Each study compares conditions that differ in one respect, with paired bootstrap 95\% intervals over scene groups on navhard and over frames on navtest. Analyses whose rules predate their runs are the control comparisons, the step-by-step build and the rare-scene comparison. The split into hard scenes is exploratory. The navtest numbers of Tables~\ref{tab:steps} and \ref{tab:main} come from the official one-stage scorer, and the slices of Table~\ref{tab:hard} from our own paired replay. At test time the system tracks candidate paths with the benchmark's controller and bicycle model to compute the fact measures and the memory's motion profiles, but no component accepts or rejects an output by a benchmark check.

\section{Results}
We ask four questions in order. Are the stated facts right (Sec.~\ref{sec:reasoner})? Does a fact change the driving, and can we tell why (Sec.~\ref{sec:notwork})? How strong is the system around the frozen planner (Sec.~\ref{sec:steps})? Where does the free-distance bound help (Sec.~\ref{sec:hard})? The first two use the reasoner before its repair, and the last two use the repaired one.

\subsection{Are the stated facts right?}\label{sec:reasoner}
\begin{table}[!h]
\centering
\caption{\textbf{Are the stated facts right?} The fact sheet of the reasoner before its repair against map and agent labels at $t{=}0$ on 1{,}600 held-out scenes, 800 real (navtrain, recorded camera images) and 800 rendered (SimScale, synthesized camera images). The base model is Cosmos-Reason2-8B with the same prompt and token budget. Unreadable answers count as wrong for recall and accuracy, precision is over readable answers, and a dash means nothing was stated to score. Free distance is scored on scenes whose label states a stop.}
\label{tab:right}
\setlength{\tabcolsep}{3pt}\renewcommand{\arraystretch}{0.94}
\resizebox{\columnwidth}{!}{%
\begin{tabular}{lcccc}
\toprule
 & \multicolumn{2}{c}{Before fine-tuning} & \multicolumn{2}{c}{Fine-tuned}\\
\cmidrule(lr){2-3}\cmidrule(lr){4-5}
Fact & real & rendered & real & rendered\\
\midrule
Answers that parse as a fact sheet & 7\% & 27\% & 100\% & 100\%\\
$d_{\mathrm{free}}$ median / mean error (m) & 8.3 / 16.8 & 6.1 / 12.3 & 1.0 / 11.9 & 1.0 / 6.2\\
$d_{\mathrm{free}}$ within 2\,m & 1\% & 6\% & 72\% & 79\%\\
$d_{\mathrm{free}}$ a stop stated as clear & 3\% & 7\% & 20\% & 11\%\\
Vehicle ahead recall / precision & 0.05 / 0.85 & 0.27 / 0.89 & 0.86 / 0.88 & 0.92 / 0.90\\
Vehicle ahead gap $g$, median error (m) & 3.1 & 2.5 & 0.5 & 0.5\\
Red light recall / precision & 0.03 / 0.38 & 0.35 / 0.87 & 0.65 / 0.77 & 0.89 / 0.87\\
Ego lane accuracy & 0.06 & 0.22 & 0.84 & 0.81\\
Route widths $w$ within tolerance & 0\% & 0\% & 60\% & 65\%\\
Road users within 20\,m recall / precision & 0.00 / 0.00 & 0.00 / 0.00 & 0.87 / 0.84 & 0.91 / 0.92\\
Road-user position median error (m) & -- & -- & 0.5 & 0.5\\
\bottomrule
\end{tabular}}
\end{table}
Fine-tuning turns a model that can hardly write a fact sheet into one that always does (Table~\ref{tab:right} and Fig.~\ref{fig:facts}). The free distance is within 2\,m of the map in about three scenes of four, and the vehicle ahead is found in most cases. Errors remain on cases that matter: the reasoner still calls a real stop clear in one real scene in five and misses about a third of the red lights on real frames.

The repair, a last round of fine-tuning on the most frequent errors, raised the accuracy of light, stop-line and distant-vehicle facts. The navhard gain of the five-clause fact check did not rise: 1.2 points after the repair against 2.4 before it. Better facts on average did not raise the gain of the fact check.

\subsection{Does a fact change the driving, and can we tell why?}\label{sec:controls}\label{sec:notwork}
The fact check changes the driving and scores above no fact check, but we cannot yet show that fact content causes the improvement. Every change of the five-clause fact check on the frozen planner names the clause behind it (Table~\ref{tab:iface}). On navhard it scores 2.4 EPDMS points above no fact check and 3.5 above shuffled facts. Over the matched control, which changes the same frames to the next-best path without any fact, the margin is 2.0 and its interval, $[-0.1, +4.2]$, includes zero. The evidence therefore supports a stated reason for every change, but not yet that the facts raise the score.

\begin{figure*}[t]
\centering
\includegraphics[width=0.89\textwidth]{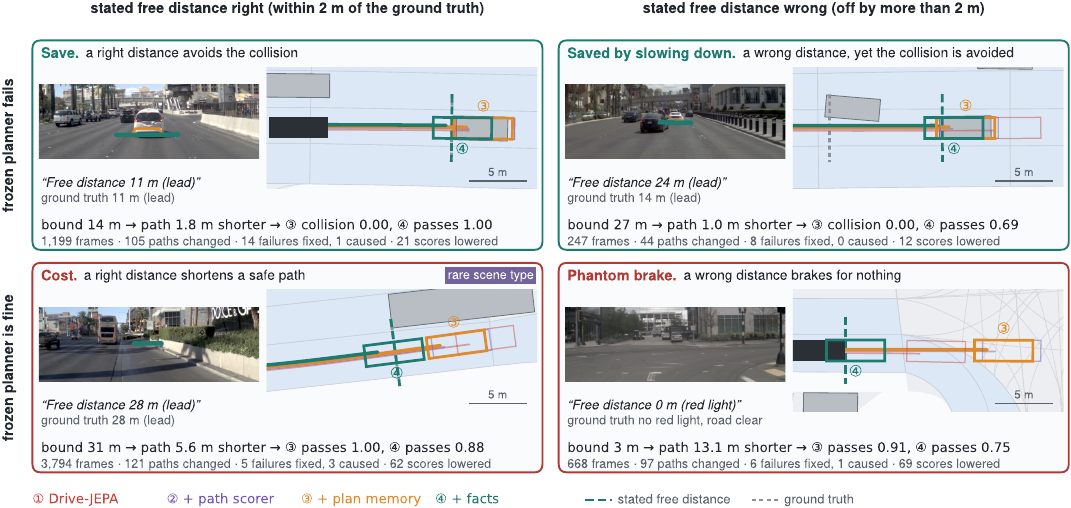}
\caption{\textbf{Free-distance bound on four navhard frames.} Rows show whether frozen Drive-JEPA fails a hard check; columns, whether stated free distance is within 2\,m of ground truth. Each cell shows the $t{=}0$ camera, road users, end-of-path map for \circled{1} frozen planner, \circled{2} scorer, \circled{3} memory and \circled{4} facts, the simulator's frame scores (0--1) for \circled{3}/\circled{4}, and frequency among 5{,}908 readable frames (5{,}912 total). Hardness is per frame here, per group in Table~\ref{tab:hard}. Examples are median real frames of each pattern; each upper cell has one eligible frame, and the lower-right has a clear road.}
\label{fig:steps}
\end{figure*}
\begin{table}[!ht]
\centering
\caption{\textbf{One fact sheet, two interfaces.} navhard EPDMS on the frozen planner, with the reasoner before its repair. Each change is the paired difference from its reference over 225 scene groups, with its 95\% interval. Shuffled facts come from another scene. Exact facts come from the map and the annotated agents and are a diagnostic, never a method.}
\label{tab:iface}
\footnotesize
\setlength{\tabcolsep}{3pt}
\begin{tabular}{lccl}
\toprule
Facts used as & EPDMS & Change [95\%] & Against\\
\midrule
none, frozen planner & 41.2 & & \\
five-clause fact check, & 43.6 & $+$2.4 [$+$0.1, $+$4.7] & no fact check\\
\quad reasoner's facts & & $+$3.5 [$+$1.0, $+$5.9] & shuffled facts\\
 & & $+$2.0 [$-$0.1, $+$4.2] & matched control\\
five-clause fact check, & 45.6 & $+$4.4 [$+$2.4, $+$6.4] & no fact check\\
\quad exact facts (oracle) & & $+$4.4 [$+$2.4, $+$6.5] & shuffled facts\\
 & & $+$3.6 [$+$1.7, $+$5.6] & matched control\\
\midrule
input to the re-ranker & 35.5 & $-$1.1 [$-$2.9, $+$0.6] & no facts (36.7)\\
input to the retrained head & 39.4 & $-$1.4 [$-$3.4, $+$0.6] & no facts (40.8)\\
\bottomrule
\end{tabular}
\end{table}

A stated reason is not always a correct one. With the reasoner before its repair, the fact check changes 1{,}221 navhard frames, and the simulator confirms the stated fact in 40\% of them. We call a change confirmed when the simulator's matching test of that fact fails on the path the fact check removed. An auditor can review every change, although some stated reasons will be wrong.

The same facts gave no gain as inputs to learned components (Table~\ref{tab:iface}). The re-ranker changes its choice in more than a third of navhard frames when we remove the facts, so it reads them. Yet it scores 35.5 with the facts against 36.7 without, and its chosen path satisfies the fact sheet in only about half of the frames. The retrained head obeys exact facts on training scenes, yet with the reasoner's facts scores 39.4 against 40.8 without. It was trained on exact map facts and tested with the reasoner's facts. Both intervals include zero, so these models did not benefit, and we do not show a loss. All four learned versions score below the frozen planner's 41.2, so only facts versus no facts counts. These are weak baselines, and a stronger learned design may use the facts better. The two interfaces gave different outcomes, though each arm is judged against its own reference.

\para{How much could exact facts help?} As a diagnostic, we replace the reasoner's facts by exact facts computed from the map and annotated road users at $t{=}0$. This oracle is never a method. With exact facts, the five-clause fact check scores 45.6: 4.4 points above no fact check and 3.6 above its matched control, with both intervals above zero (Table~\ref{tab:iface}). Each matched control changes the same frames as its own fact check, so the controls differ. Exact facts are 2.0 points above the reasoner's facts, but that interval includes zero [$-$0.1, $+$4.2]. Right facts could therefore carry value, but this sample cannot show how much the reasoner's errors cost.

\begin{table}[!t]
\centering
\caption{\textbf{Step by step.} Each row adds one part to the frozen planner. Stage~1 scores real frames and stage~2 their simulated follow-up states, and Fail is the share of frames that fail a hard check (stage~1 / stage~2). Small numbers show the change from the row above, bold if significant at 95\%. Changes are computed from unrounded scores. Step~3 is the base system and step~4 the full system. Navtest covers all 12{,}146 scenes. The navhard momentum row is our rebuild, with two earlier frames of history.}
\label{tab:steps}
\setlength{\tabcolsep}{1.35pt}\renewcommand{\arraystretch}{1.15}
\resizebox{\columnwidth}{!}{%
\begin{tabular}{lccccc}
\toprule
 & \multicolumn{4}{c}{navhard} & navtest\\
\cmidrule(lr){2-5}\cmidrule(lr){6-6}
Step & Stage 1 & Stage 2 & EPDMS & Fail (\%) & EPDMS\\
\midrule
\circled{1} Drive-JEPA, frozen & 78.4 & 52.8 & 41.2 & 8.7 / 28.6 & 87.9\\
\circled{2} + path scorer & 80.5\,\dd{+2.2} & 62.2\,\ddb{+9.4} & 50.7\,\ddb{+9.5} & 5.6 / 18.0 & 88.1\,\dd{+0.2}\\
\rowcolor{ours}\circled{3} + plan memory & 80.4\,\dd{$-$0.1} & 64.2\,\ddb{+1.9} & 51.8\,\dd{+1.2} & 5.8 / 18.1 & 90.2\,\ddb{+2.1}\\
\rowcolor{ours}\circled{4} + facts & 80.2\,\dd{$-$0.2} & 64.9\,\dd{+0.7} & 52.2\,\dd{+0.3} & 5.1 / 16.9 & 88.2\,\ddb{$-$2.0}\\
\midrule
\shortstack[l]{\circled{1} with its momentum step\\(test-time check)} & 82.6 & 52.8 & 43.5 & 8.7 / 30.9 & 90.8\\
\circled{4} with shuffled facts & 76.9 & 62.6 & 48.6 & 4.4 / 17.5 & 83.9\\
\bottomrule
\end{tabular}}
\end{table}

\begin{table*}[!t]
\centering
\caption{\textbf{Comparison on NAVSIM v2.} navtest reports one-stage EPDMS and its sub-metrics on all 12{,}146 scenes, and navhard the two-stage EPDMS, where stage~1 scores real frames and stage~2 the simulated start states that follow. Published numbers come from each method's paper. $^\ddagger$ marks scores before the devkit fix of 29 Sep 2025, which raised EPDMS by 2 to 7 points, $^\S$ an unstated version, $^{*}$ the navhard leaderboard~\cite{cao2025pseudosim} and $^\dagger$ privileged ground-truth inputs. Our rows keep Drive-JEPA frozen and use the official scorer locally after the fix (the frozen planner's 41.2 matches the official scorer), so they are not leaderboard submissions. Comfort between plans, EC, covers 10{,}040 scenes with a previous frame. With its momentum step, frozen Drive-JEPA has EC 76.9 and EPDMS 90.8 (Table~\ref{tab:steps}). Published navtest scores use other devkit versions, so only our rows are compared. Inputs are cameras (C) or ground truth (GT). Bold marks the best of our rows. The shaded rows are our two systems, and we recommend the base system for driving score because it is best of ours on navtest. The full system shows what the facts add, and its 0.3 navhard gain is within noise.}
\label{tab:main}

\setlength{\tabcolsep}{3.2pt}\renewcommand{\arraystretch}{0.86}
\resizebox{\textwidth}{!}{%
\begin{tabular}{llc cccccccccc ccc}
\toprule
 & & & \multicolumn{10}{c}{navtest (NAVSIM v2, one stage)} & \multicolumn{3}{c}{navhard (two stage)}\\
\cmidrule(lr){4-13}\cmidrule(lr){14-16}
Method & Venue & Input & NC & DAC & DDC & TLC & EP & TTC & LK & HC & EC & EPDMS & Stage 1 & Stage 2 & EPDMS\\
\midrule
\multicolumn{16}{l}{\textit{Rule-based}}\\
PDM-Closed$^\dagger$~\cite{dauner2023pdm} & CoRL'23 & GT & -- & -- & -- & -- & -- & -- & -- & -- & -- & -- & -- & -- & 56.6\\
\midrule
\multicolumn{16}{l}{\textit{End-to-end and trajectory scoring}}\\
Hydra-MDP++~\cite{li2025hydramdppp} & arXiv'25 & C & 98.5 & 98.5 & 99.5 & 99.7 & 87.4 & 97.9 & 95.8 & 98.2 & 75.7 & 85.6$^\ddagger$ & -- & -- & --\\
ZTRS~\cite{li2025ztrs} & arXiv'25 & C & 98.2 & 99.1 & 99.7 & 99.8 & 86.9 & 97.5 & 96.6 & 98.2 & 78.2 & 86.2$^\ddagger$ & -- & -- & 48.1$^{*}$\\
DrivoR~\cite{kirby2026drivor} & CVPR'26 & C & -- & -- & -- & -- & -- & -- & -- & -- & -- & -- & -- & -- & 48.3\\
DriveZero~\cite{he2026drivezero} & arXiv'26 & C & -- & -- & -- & -- & -- & -- & -- & -- & -- & -- & 84.1 & 60.8 & 51.5\\
DrivoR + SimScale~\cite{kirby2026drivor,tian2026simscale} & CVPR'26 & C & -- & -- & -- & -- & -- & -- & -- & -- & -- & -- & -- & -- & 54.6\\
TOAD~\cite{xu2026toad} & arXiv'26 & C & -- & -- & -- & -- & -- & -- & -- & -- & -- & -- & -- & -- & 56.3\\
Drive-JEPA~\cite{wang2026drivejepa} & arXiv'26 & C & 98.4 & 98.6 & 99.1 & 99.8 & 88.4 & 97.8 & 97.6 & 97.9 & 84.8 & 87.8$^\ddagger$ & -- & -- & --\\
\midrule
\multicolumn{16}{l}{\textit{Vision-language-action}}\\
ReCogDrive~\cite{li2026recogdrive} & ICLR'26 & C & 98.3 & 95.2 & 99.5 & 99.8 & 87.1 & 97.5 & 96.6 & 98.3 & 86.5 & 83.6$^\ddagger$ & -- & -- & --\\
ELF-VLA~\cite{luo2026elfvla} & CVPR'26 & C & 98.9 & 98.1 & 99.4 & 99.8 & 88.5 & 98.4 & 96.9 & 98.3 & 87.2 & 87.1$^\S$ & -- & -- & --\\
RefineDrive~\cite{sun2026refinedrive} & arXiv'26 & C & 98.7 & 98.2 & 98.1 & 99.7 & 90.5 & 98.2 & 94.4 & 97.8 & 82.7 & 89.4 & -- & -- & --\\
HiST-VLA~\cite{wang2026histvla} & arXiv'26 & C &
99.6 & 99.1 & 99.7 & 99.9 & 89.2 & 99.4 & 98.9 & 98.4 & 66.0 & 88.6 &
-- & -- & 50.9\\
\midrule
\multicolumn{16}{l}{\textit{World-model augmented}}\\
Discrete-WAM~\cite{yao2026discretewam} & arXiv'26 & C & 98.5 & 98.2 & 99.7 & 99.8 & 90.5 & 97.9 & 97.2 & 98.3 & 78.1 & 90.4 & -- & -- & --\\
DriveFuture~\cite{hong2026drivefuture} & arXiv'26 & C & 98.8 & 99.1 & 99.6 & 99.9 & 86.6 & 98.4 & 96.4 & 98.3 & 74.8 & 89.9 & -- & -- & 55.5$^{*}$\\
SUV~\cite{yuan2026suv} & arXiv'26 & C & 99.1 & 97.8 & 99.7 & 99.8 & 87.8 & 98.7 & 98.1 & 98.4 & 88.3 & 91.0 & 82.3 & 43.9 & 36.9\\
\midrule
\multicolumn{16}{l}{\textit{Frozen Drive-JEPA (ours, local official scorer)}}\\
Drive-JEPA, frozen (our run) & -- & C & 98.5 & 98.6 & 99.7 & 99.8 & 92.1 & 97.9 & 97.6 & 97.8 & 44.6 & 87.9 & 78.4 & 52.8 & 41.2\\
+ path scorer & -- & C & 99.1 & 99.0 & 99.7 & 99.9 & 90.0 & 98.5 & 97.2 & 98.1 & 44.2 & 88.1 & 80.5 & 62.2 & 50.7\\
\rowcolor{ours}+ path scorer, plan memory (base system) & -- & C & 98.9 & 99.0 & 99.8 & 99.9 & 87.4 & 98.1 & 96.8 & 97.3 & 75.5 & \B{90.2} & 80.4 & 64.2 & 51.8\\
\rowcolor{ours}+ path scorer, plan memory, facts (full system) & -- & C & 98.6 & 99.0 & 99.8 & 99.9 & 84.7 & 97.6 & 96.7 & 95.4 & 70.3 & 88.2 & 80.2 & 64.9 & \B{52.2}\\
\bottomrule
\end{tabular}}
\end{table*}

\subsection{How strong is the system built around the frozen planner?}\label{sec:steps}\label{sec:system}

We add the parts to the frozen planner one at a time (Table~\ref{tab:steps}). The path scorer does most of the work, nearly all in stage~2, the simulated follow-up states reached by the frozen planner's own plans and absent from training logs. There it cuts hard-check failures from 28.6\% to 18.0\%, and navhard rises from 41.2 to 50.7. The plan memory keeps consecutive plans consistent, lifting navtest through its comfort term and navhard by a smaller amount. Drive-JEPA's released momentum step reaches 90.8 on navtest, but applies the benchmark's comfort check at test time. The base system recovers 78\% of that gain without applying the check.

The facts enter last, as the free-distance bound. They move the average score very little, up 0.3 on navhard and down 2.0 on navtest. This is not because the planner ignores them: facts from another scene score 3.6 points lower on navhard, so the bound reacts to what the facts say and wrong facts hurt. This does not show that right facts help. Where they help is the question of Sec.~\ref{sec:hard}.

As a diagnostic, exact free distances from the map and annotated agents in place of the reasoner's raise the full system to 52.9 on navhard. That is 1.0 above the base system [$-$0.1, $+$2.4], 0.7 above the reasoner's facts [$-$0.2, $+$1.7], and 4.2 above the same distances shuffled between frames [$+$2.4, $+$6.1]. Even with exact distances, the bound has little average headroom.

Table~\ref{tab:main} places our systems next to published planners. The frozen Drive-JEPA is below most published navhard scores. The full system reaches 52.2, inside the published camera-based range of 36.9 to 56.3 and below its best. Most of the gain comes from parts that never read the facts. Facts exist for the 10{,}795 navtest scenes with camera files; the other 1{,}351 keep the base system's choice.

\subsection{Where does the free-distance bound help?}\label{sec:hard}\label{sec:where}
The last question is where the free-distance bound pays. We split the scenes by hardness and rarity, as defined in the introduction (Table~\ref{tab:hard}). The two splits give different answers.

\begin{table}[!t]
\centering
\caption{\textbf{Hard, not rare.} Point changes of the full system, by slice. Facts vs none is the full system minus the base system, and facts vs shuffled is the full system minus the same system with shuffled facts. A hard navhard group has a stage~1 frame where the frozen planner fails a hard check. Rare groups have the least frequent scene types among 30{,}944 training scenes. Right is, for the repaired reasoner, the share of current frames whose stated free distance is within 2\,m of the map. Navtest slices come from our own paired replay on 10{,}795 scenes with camera files, which scores the base and full systems at 89.3 and 87.4, not 90.2 and 88.2 as in Table~\ref{tab:steps}. 95\% intervals.}
\label{tab:hard}
\footnotesize
\setlength{\tabcolsep}{2pt}\renewcommand{\arraystretch}{0.94}
\resizebox{\columnwidth}{!}{%
\begin{tabular}{lrccc}
\toprule
Slice & $n$ & Facts vs none & Facts vs shuffled & Right\\
\midrule
navhard, all groups & 225 & $+$0.3 [$-$0.6, $+$1.4] & $+$3.6 [$+$1.8, $+$5.4] & 81\%\\
\quad hard & 29 & $+$3.4 [$-$1.0, $+$8.7] & 0.0 [$-$4.4, $+$4.5] & 76\%\\
\quad not hard & 196 & $-$0.1 [$-$0.9, $+$0.7] & $+$4.1 [$+$2.2, $+$6.1] & 82\%\\
\quad rare scene type & 49 & $+$0.3 [$-$1.8, $+$2.9] & $+$4.1 [$+$0.9, $+$7.4] & \B{51\%}\\
\quad common scene type & 176 & $+$0.3 [$-$0.7, $+$1.4] & $+$3.4 [$+$1.4, $+$5.5] & 90\%\\
\midrule
navtest, all scenes & 10{,}795 & $-$1.9 [$-$2.1, $-$1.7] & $+$4.7 [$+$4.4, $+$5.0] & --\\
\quad hard & 344 & $+$3.7 [$+$1.0, $+$6.4] & $+$0.6 [$-$2.7, $+$3.8] & --\\
\quad not hard & 10{,}451 & $-$2.1 [$-$2.3, $-$1.9] & $+$4.8 [$+$4.5, $+$5.1] & --\\
\bottomrule
\end{tabular}}
\vspace{-4mm}
\end{table}
\para{Hard~frames.} In this exploratory split the bound helps where the planner fails. On navtest it gains 3.7 points on the 344 hard scenes and costs 2.1 on all others, because a stated free distance can shorten a path the planner would otherwise drive safely. Hard scenes are only about 3\% of navtest, while about 29\% of simulated navhard frames are hard, so benchmark composition matters. On hard groups of both benchmarks, the reasoner's facts score about the same as shuffled ones. We therefore find no sign that their content matters on hard scenes, although the intervals are wide. As an exploratory oracle diagnostic, exact distances beat shuffled exact distances by 9.4 points on the 29 hard navhard groups [$+$2.1, $+$18.3].

\para{Rare~scenes.} We could not detect that rarity predicts where the bound helps. Rare and common scene types gain a similar small amount on navhard, a difference of 0.0 with a wide 95\% interval [$-$2.4, $+$2.7]. The reasoner is the weak link on rare scenes: its free distance is within 2\,m of the map in about half of rare groups against nine in ten common ones, and a bound can only be as good as the distance it is given. Fig.~\ref{fig:steps} puts both findings in one picture. Where the frozen planner fails, the bound fixes failures with a right distance and also with a wrong one, suggesting that stopping earlier is what helps there. Where the planner is fine, it costs progress, rarely with a right distance and far more often with a wrong one, such as a red light stated where the ground truth has none.

\section{Discussion and Limitations}\label{sec:discussion}

\para{What the results mean.} Our evidence supports a checkable reason for each change the fact check makes, not that the reasoner's fact content causes the gain. With exact facts, the margin over the matched control is clear of zero, so right facts could carry value, although their gain over the reasoner's facts includes zero.

\para{Implications for safety cases.} A safety case collects a reason for each design choice that a reviewer can check~\cite{ul4600}. The two learned components give no such reason because we cannot say which fact moved a choice. The fact check names the clause that removed or shortened each path, letting a reviewer compare the stated fact with the map. This audit trail does not make the reason correct: a missed fact produces no change, and before repair the simulator confirms the stated fact in only 40\% of changes. It therefore supports, but does not replace, evidence that each fact is right.

\para{Limitations.} Evaluation covers one frozen planner, one VLM family and NAVSIM v2; transfer to other planners, datasets and on-road driving remains untested. The two learned components are weak baselines, all below the frozen planner, so we do not claim that no learned design can use the facts. Exact-fact runs are an oracle diagnostic on navhard only, at fact-sheet resolution and with clause tolerances set for the reasoner's facts. The fact sheet covers predefined attributes, so unrepresented hazards cannot affect candidate selection. The hard slice is retrospective and diagnostic, not a deployable trigger. The full system also retains a comfort gap: only about half of consecutive plans pass the benchmark's between-plan comfort test.

\section{Conclusion and Future Work}

We asked where facts stated by a VLM must enter a frozen driving planner to matter. As inputs to the two learned components we tried, they changed choices without improving driving. As a fact check on candidate paths, they gave every change a stated clause, yet only exact facts put the matched-control margin clear of zero. A path scorer and plan memory, which read no facts, supply most of the gain. The fact check makes reasoning influence checkable for a safety argument. Future work should improve fact reliability on rare scenes and test other planners and benchmarks.

\bibliographystyle{IEEEtran}
\bibliography{refs}

\end{document}